\documentclass[sn-mathphys,Numbered]{sn-jnl}

\usepackage{graphicx}%
\usepackage{multirow}%
\usepackage{amsmath,amssymb,amsfonts}%
\usepackage{amsthm}%
\usepackage{mathrsfs}%
\usepackage[title]{appendix}%
\usepackage{xcolor}%
\usepackage{textcomp}%
\usepackage{manyfoot}%
\usepackage{booktabs}%
\usepackage{algorithm}%
\usepackage{algorithmicx}%
\usepackage{algpseudocode}%
\usepackage{listings}%
\usepackage{pdfpages}

\theoremstyle{thmstyleone}%
\newtheorem{theorem}{Theorem}
\theoremstyle{thmstyletwo}%

\theoremstyle{thmstylethree}%

\begin{document}

\title[A.H.D]{A Hamiltonian driven Geometric Construction of Neural Networks via the Lognormal
family, Application to Financial Fraud Detection and to Network
Security}

\author*[1]{\fnm{} \sur{Prosper Rosaire Mama  Assandje }}\email{mamarosaire@facsciences-uy1.cm}
\author[1]{\fnm{} \sur{Landry Foka Marius}}\email{lamdrymarius@mail.com}
\equalcont{These authors contributed equally to this work.}
\author[1]{\fnm{} \sur{Arnaud Gires Fobasso Tchinda}}\email{fobass1989@gmail.com}
\equalcont{These authors contributed equally to this work.}
\author[2]{\fnm{} \sur{Fr$\acute{e}$d$\acute{e}$ric Barbaresco }}\email{frederic.barbaresco@thalesgroup.com}
\equalcont{These authors contributed equally to this work.}
\author[1]{\fnm{} \sur{St$\acute{e}$phane R.  Gael Ekodeck}}\email{stephane-gael.ekodeck@facsciences-uy1.cm}
\equalcont{These authors contributed equally to this work.}
\author[1]{\fnm{} \sur{Serge Alain Ebele}}\email{serge-alain.ebele@facsciences-uy1.cm}
\equalcont{These authors contributed equally to this work.}

\affil*[1]{\orgdiv{Department of Mathematics}, \orgname{University
of Yaounde 1}, \orgaddress{\street{usrectorat@.univ-yaounde1.cm},
\city{yaounde}, \postcode{812}, \state{Center}, \country{Cameroon}}}
 \affil[2]{\orgdiv{Department of Mathematics}, \orgname{Thales Land \& Air Systems(
Industry ) }, \orgaddress{\street{https://www.thalesgroup.com/en},
\city{2 Avenue Gay Lussac}, \postcode{ CS90502 78990 Elancourt},
\state{Region 8 (Africa, Europe, Middle East)}, \country{France}}}


\abstract{We presents a method for constructing neural networks
intrinsically on statistical manifolds via the lognormal
distribution. We demonstrate this approach by formulating a neural
network architecture directly on statistical manifold. The
construction is driven by the Hamiltonian system that is equivalent
to the gradient flow on this manifold. We define the network's input
values using the coordinate system of this Hamiltonian dynamics,
naturally embedded in the Poincar$\acute{e}$ disk. The core of our
contribution lies in the derivation of the network's components from
geometric principles: the rotation component of the synaptic weight
matrix is determined by the Lie group action of $SU(1,1)$ on the
disk, while the activation function emerges from the symplectic
structure of the system. We subsequently obtain the complete weight
matrix, including its translation vector, and the resulting output
values.}
\keywords{   Hamiltonian system, Neural network, activation
function.}


\pacs[MSC Classification]{53D05, 51H20, 51H25 , 53Z05, 53Z30
,62E10,20P05}
 \maketitle
\section{Introduction}\label{sec1}

It is well know that  artificial neural network is a mathematical
model of nervous system. It is consisting of layers of
interconnected nodes, that transform input data into output through
weighted connection and activation functions. As a mathematical
model of biological nervous system, artificial neural networks plays
an important role in machine learning, in the theory of
reconstitution and form recognition. According to Amari's
\cite{amari:95}, this modelization of biological neurones networks
can be make using appropriate statistical manifold's. More
precisely, since neural networks is compose of large number of
neurones connected with each other; the set of connection weights
$\Theta=\left\{\theta^{j}=\left(\theta^{j}_{1},\dots,
\theta^{j}_{n}\right)\right\}_{j\in\Lambda}$ where $\theta^{j}_{i}$
modelized the nerve flows from neuron $i$ to neuron $j$ through
nerve $ij$  define a structure of $n-$dimensional manifold with
coordinates systems $\Theta$. It is usually call neural manifold
\cite{amari:95}. The geometry of neural manifold help us to
understanding the total capability of a class of network
\cite{amari:95}. The modeling of artificial neural networks has
traditionally relied on Euclidean spaces, thus limiting their
ability to capture the complexity of structured data on statistical
manifolds. Amari's seminal work paved the way for a geometric
approach to neural networks, where the manifold of synaptic
parameters is endowed with a Riemannian structure. However, few
studies have explored the explicit construction of neural networks
on specific statistical manifolds, such as the lognormal manifold,
which is ubiquitous in finance, biology, and signal processing. This
article follows this approach by proposing a systematic method for
building a neural network on the lognormal manifold by exploiting
its Hamiltonian and geometric structure. Using the
Poincar$\acute{e}$ disk and the action of the $SU(1,1)$ group, we
show how information processing systems can be naturalized on
non-Euclidean spaces, thus offering a new perspective for geometric
machine learning. We recall that information geometry has been
applied successfully to information theory, statistical sciences,
and systems theory. In $1991$, Amari \cite{amari:95}, propose its
first application to neural networks. According to McCulloch and
Pitts in \cite{all:07}, the theory of artificial neural networks was
introduced around $1940$. In $1946$, at the New York University,
Shannon introduced the concept of information theory \cite{all:07}.
He shows that Wiener's cybernetics and the transistor where the
begining of machine intelligence and information theory. In $1950$,
Frank Rosenblatt's invented the perceptron, the forerunner of
multilayer neural networks. It is shown in \cite{souriau:21} that
the trajectories of the drones are encoded by the time series of the
Euclidean special group $SE(3)$ provided by an invariant extended
Kalman filter cracker radar based on the local Frenet Seret model.
Riken introduces in \cite{souriau:21}, a natural gradient based on
the Fisher information matrix. Jean Marie Souriau established in
\cite{souriau:21}, a theorem saying that every symplectic manifold
is a Co-adjoint orbit manifold based on the homogeneous symplectic
Lie group classification. It is prove in \cite{souriau:21} that
Jean-Marie Souriau has proposed a model of machine learning on
homogeneous symplectic manifold of Lie group co-adjoint orbits. In
\cite{mama:23}, it was shown that the gradient system defined on the
lognormal manifold is equivalent to the Hamiltonian system and is
completely integrable. In \cite{mauro1:25}, it is shown that
convergence of dynamic feedback neural networks (NNs), as the
Cohen-Grossberg, Hopfield and cellular NNs, has been for a long time
a workhorse of NN theory. Indeed, convergence in the presence of
multiple stable equilibrium points (EPs) is crucial to implement
content addressable memories and solve several other signal
processing tasks in real time. There are two typical ways to use a
convergent NN, i.e.: a) let the activations evolve while maintaining
fixed weights and inputs (activation dynamics) or b) adapt the
weights while maintaining fixed activations (weight dynamics).  The
weight-activation dynamics is of importance also because it is more
plausible than the other two types for modeling neural systems. In
\cite{mauro2:24}, it is to investigate convergence for the classic
Brain State in a Box (BSB) NN model and some of its relevant
generalizations named Brain State in a Convex Body (BSCB). M Di
Marco  \cite{mauro3:23}, studies via the Lyapunov method complete
stability (CS), i.e., convergence of trajectories in the presence of
multiple equilibrium points (EPs), for delayed NNs with Stanford
memristors. Given a lognormal statistical manifold, can it be seen
as a neural manifold of a given neural network and how can we
structure the lognormal statistical manifold as a neural manifold,
and deduce the components of an artificial neural network, including
the input-output functions, the weight matrix, and the activation
function, using Hamiltonian geometry and the Lie group formalism?
This leads us to sub-questions: How does the transitive action of
the $SU(1,1)$ group on the Poincar$\acute{e}$ disk allow us to
define the network's inputs and outputs? In which case, what are the
input functions, output functions and the weight matrix of this
network? The dynamics in the system is governed by the Hamiltonian
formalism, which in turn is related to the gradient system on the
lognormal manifold. To construct a structure of the neural manifold
from a lognormal manifold, we will identify the input and output
values and the synaptic matrix from the activation function to be
constructed. We then propose a method to determine the input
functions, the weight matrix and the output functions for a neural
network on statistical manifolds. We show that the Hamiltonian of
the gradient system defined on the unit disk of Poincar$\acute{e}$
of the lognormal family is given by
\[\mathcal{H}=-\left(P^{2}+Q^{2}\right)\cos(
\beta)\sin( \beta),\] where $P=\frac{2\theta_{2}}{\theta_{1}},\;
Q=\frac{\theta^{2}_{1}-2\theta_{2}}{4\theta^{2}_{2}}$,
$\theta_{1}=\frac{\mu}{\sigma^{2}},\;
\theta_{2}=-\frac{1}{2\sigma^{2}}$, $\cos(
\beta)=\frac{P}{\sqrt{P^{2}+Q^{2}}}$, and $\sin(
\beta)=\frac{Q}{\sqrt{P^{2}+Q^{2}}}$. The integrable Hamiltonian
system  has trajectories that are closed curves (ellipses) in the
phase space $(P,Q)$. In classical mechanics, such a system can often
be compactified or visualized more naturally on a projective or
hyperbolic space. The Poincar$\acute{e}$ disk $D$ is a model of
hyperbolic geometry. The group $SU(1,1)$ is the group of isometries
that preserve this geometry. The action of $SU(1,1)$ on $D$ then
describes the dynamical evolution of the system. This is a way of
modeling neural dynamics in a space of constant curvature. We show
that the neural information processing systems of the lognormal
family in the complex plane is  given by the following equation
\[g_{1}.z=\frac{\frac{1}{2} \cosh( \cos( \beta))\cos( \beta)+\sinh( \cos( \beta)) +
\frac{1}{2}i\cosh( \cos( \beta))\sin( \beta) }{\frac{1}{2} \sinh(
\cos( \beta))\cos( \beta)+\cosh( \cos( \beta)) + \frac{1}{2}i\sinh(
\cos( \beta))\sin( \beta)},\]
\[g_{2}.z=\frac{\frac{1}{2} \cosh( \sin( \beta))\sin(
\beta)+\sinh(\sin( \beta)) +\frac{1}{2}i\cosh( \sin( \beta))\sin(
\beta) }{\frac{1}{2} \sinh( \sin( \beta))\sin( \beta)+\cosh(\sin(
\beta)) +\frac{1}{2}i\sinh( \sin( \beta))\sin( \beta)},\]
\[
g^{-1}_{1}.z=\frac{\frac{1}{2}\cosh\left(\cos(
\beta)\right)\cos(\beta)-\sinh\left(\cos(
\beta)\right)+\frac{1}{2}i\sin( \beta)
}{-\frac{1}{2}\sinh\left(\cos(
\beta)\right)\cos(\beta)+\cosh\left(\cos(
\beta)\right)+i\frac{1}{2}\sin( \beta)},\] and
\[
g^{-1}_{2}.z=\frac{\frac{1}{2}\cosh\left(\sin(
\beta)\right)\cos(\beta)-\sinh\left(\sin(
\beta)\right)+\frac{1}{2}i\sin( \beta)
}{-\frac{1}{2}\sinh\left(\sin(
\beta)\right)\cos(\beta)+\cosh\left(\sin(
\beta)\right)+i\frac{1}{2}\sin( \beta)}.\] We show that the neural
network system based on the lognormal manifold  is given by $\left(
                           \begin{array}{c}
                             Z' \\
                             1 \\
                           \end{array}
                         \right)=\left(
                                   \begin{array}{cc}
                                     \Omega & \vec{t} \\
                                     0 & 1\\
                                   \end{array}
                                 \right)\left(
                           \begin{array}{c}
                             Z \\
                             1 \\
                           \end{array}
                         \right)
$. With $Z=\left(
                           \begin{array}{c}
                             \frac{1}{2}\cos( \beta) \\
                            \frac{1}{2}\sin( \beta) \\
                           \end{array}
                         \right),$ $Z'=\left(
                           \begin{array}{c}
                             e^{\left(\frac{1}{2}\cos(
\beta)\right)}\cos(\sin( \beta)) \\
                             e^{\left(\frac{1}{2}\cos(
\beta)\right)}\sin(\sin( \beta)) \\
                           \end{array}
                         \right)$,\\$\;$ $\Omega=\left(
             \begin{array}{cc}
               \cos( \beta) & -\sin( \beta) \\
               \sin( \beta) & \cos( \beta) \\
             \end{array}
           \right)$ is a rotation matrix $(2\times2)$. It represents
            a linear transformation, similar to the weights in a standard dense layer, but with a rotation constraint.
and $\vec{t}=\left(
                 \begin{array}{c}
                   e^{\left(\frac{1}{2}\cos(
\beta)\right)}\cos(\sin( \beta))-\frac{1}{2}\cos( 2\beta)\\
                  e^{\left(\frac{1}{2}\cos(
\beta)\right)}\sin(\sin( \beta))-\cos(
                   \beta)\sin(\beta)
                   \\
                 \end{array}
               \right)\in \mathbb{R}^{2}$ is a translation  vector. It shifts the result of the linear transformation.
The proposed model is a fully connected single layer network with a
bias. Input vector: $X = \left(Z, 1\right) = \left( \frac{1}{2}\cos(
\beta), \frac{1}{2}\sin( \beta), 1 \right)$. It represents the
initial state of the information on the manifold, encoded in the
Poincar$\acute{e}$ disk. The Weight Matrix (Synaptic): $W = \left(
                                   \begin{array}{cc}
                                     \Omega & \vec{t} \\
                                     0 & 1\\
                                   \end{array}
                                 \right)$ It is an element of the Special Euclidean group $SE(2)$.
The output vector $Y = \left(Z', 1\right)$ represents the
transformed state of the information after applying the
transformation $W$. This approach allows the design of neural
network architectures whose fundamental operations (linear
combination, nonlinearity) are not simple algebraic abstractions,
but are rooted in the underlying geometry of the data. This is
particularly relevant for data from lognormal distributions, where
the parameter space is naturally a manifold. The second section
recalls the preliminary notion. In section \ref{sec55}, we present
method for constructing neural networks on statistical manifolds. In
section \ref{sec5}, we determine the neural manifold from the
lognormal statistical manifolds.

\section{Preliminaries}\label{sec2}
In this section, we review some concepts we need to know about
neural networks. According to \cite{hri:98}, a neuron is a
processing unit which have some  inputs and only one output. In
geometry \cite{all:07}, neural networks are represented by a system
that can be expressed as the product of a vector and a matrix. The
vector represents an input signal to a neural network and the matrix
represents the weight of the neural network. The output of the
neural network is the result of this product.

\subsection{Statistical manifolds}
Let $(S,\Xi, P)$ be a probability space. In \cite{boy:22}, an
$n$-dimensional statistical model for $(S, \Xi)$ is a pair $(\Theta,
p)$ in which $\Theta$ is an open subset of the Euclidean space
$\mathbb{R}^{n}$ and $p$ is a function
\begin{eqnarray*}\Theta \times \Xi\ni \left(\theta, x\right)\longmapsto p_{ \theta}(x)\in \mathbb{R}\end{eqnarray*}
satisfying the following conditions
\begin{enumerate}
\item $p_{ \theta}(x)$ is differentiable with respect to
$\theta$.
\item For all $\theta\in \Theta,\; p_{ \theta}:x\mapsto p_{ \theta}(x)$ is the density $P$ of a probability measure
on $(S, \Xi)$.
\item  If $\theta\neq \theta^{*}$ then there exists $x$ such that $p_{ \theta}(x)\neq p_{
\theta^{*}}(x)$.
\item The differentiation operator $\frac{d }{d \theta}$ and the integration
operator $\int_{\Xi}$ switch, i.e., \begin{eqnarray*}\frac{d }{d
\theta}\circ\int_{\Xi} =\int_{\Xi}\circ\frac{d }{d
\theta}\end{eqnarray*}
    \item Fisher information is always defined in an equivalent way. For all $x$ such that
 $p_{ \theta}(x)>0,\; \frac{\partial }{\partial \theta}\log p_{ \theta}(x)$ is finite.
\end{enumerate}
In \cite{lauri:87}, a statistical manifold $(S,g,\nabla)$ is a
manifold $S$ equipped with metric $g$ and connection $\nabla$ such
that
\begin{enumerate}
    \item [i)]$\nabla$ is torsion free
    \item [ii)]$\nabla$ is totally symmetric
\end{enumerate}
 Equivalently, a manifold has a statistical structure
when its $g$  conjugate  $\nabla^{*}$ of a torsion-free connection
$\nabla$ is also torsion-free.

\subsection{Gradient system and Hamiltonian system}
 In this subsection, we give some formulas and definitions of the geometrical and
statistical concepts that we will use. In \cite{ovidiu:14}, a set
$S=\left\lbrace P(x;\theta)|\theta\in \mathbb{R}^{n}\right\rbrace $
of density functions is called an exponential family if there exists
a function $\Phi:\mathbb{R}^{n}\longrightarrow\mathbb{R}$ and
$\mathbb{E}=\left\lbrace \theta\in
\mathbb{R}^{n}|\Phi(\theta)<\infty\right\rbrace $
    such that   $S=\left\lbrace P(x;\theta)|\theta\in \mathbb{E}\right\rbrace $ and
    \begin{eqnarray}
        P(x;\theta)=e^{C(x)+\theta_{i}F_{i}-\Phi(\theta)},\; \textrm{where} \; e\equiv \exp.
    \end{eqnarray}
The function $\Phi$ used above is called potential function of the
exponential family $S$.  In \cite{amari:95}, Amari prove that there
exists a pair of dual functions $\Phi$ and $\Psi$ which defined the
dual coordinate system of $S$, such that:
\begin{eqnarray}\label{dualcoord}
\left\lbrace \begin{array}{l}
\theta_{i}=\partial_{\theta_{i}}\Psi(\eta)\\
\eta_{i}=\partial_{\eta_{i}}\Phi(\theta).
\end{array}\right.
\end{eqnarray}
And they are solution of the following Legendre equation given by
\begin{eqnarray}\label{legendre}
\Phi(\theta)+\Psi(\eta)-\theta_{i}\eta_{i}=0.
\end{eqnarray}
The  function $\Psi$ is called  dual potential function of $\Phi$.
 In \cite{amari:95}, it is prove that the Fisher information of exponential family is
  defined as
    \begin{eqnarray}
        I(\theta)=\left[g_{ij}(\theta)\right]=-\left[\frac{\partial^{2}\Phi(\theta)}{\partial\theta_{i}\partial\theta_{j}}\right]_{ij}.\label{defg}
    \end{eqnarray}
 It was prove  in \cite{mama:23},  that of $S$ the gradient systems is given by
\begin{eqnarray}\dot{\overrightarrow{\theta}}&=-I(\theta)^{-1}\partial_{\theta}\Phi(\theta).\label{e2}\end{eqnarray}
The system  Eq.~(\ref{e2}) is said to be Hamiltonian if there exists
a bivector field $\barwedge$ and a smooth function $\mathcal{H}$
 such that:
\begin{equation}\dot{ \theta}(t)=\barwedge\frac{\partial \mathcal{H}}{\partial \theta}\label{e3}\end{equation}
   $\mathcal{H}$ is  called  Hamiltonian function
Furthermore $\barwedge$ satisfy $[\barwedge, \barwedge]=0$ where
$[-;-]$ denotes the Schouten Nijenhuis bracket.
\subsection{Information Geometry and Machine learning: Legendre
Structure.}\label{subsec3}
 It is well know that $SO(3)$ is a Lie group and its Lie algebra is $so(3)$.
 It is prove in \cite{Lie:10} that, $SU(1,1)=\left\{A\in SL(2,\mathbb{C})\mid A
J^{t}\bar{A}=J\right\},\; \textrm{where} \; J=\left(
                                                \begin{array}{cc}
                                                  1 & 0 \\
                                                  0 & -1 \\
                                                \end{array}
                                              \right).
$ where, $SU(1,1)$ is defined as
\begin{equation}SU(1,1)=\left\{\left(
                                \begin{array}{cc}
                                  \alpha & \xi \\
                                \bar{\xi} &  \bar{\alpha} \\
                                \end{array}
                              \right)/ |\alpha|^{2}-|\xi|^{2}=1,\; \alpha ,
                              \xi\in\mathbb{C}
\right\}\label{e21}\end{equation} with $\bar{\xi},\;\bar{\alpha}\in
\mathbb{C}$ the conjugate expression of $\alpha$,$\;$ and $\xi$.\\
Let denote the Poincar$\acute{e}$ unit Disk is given by
\begin{equation}D=\left\{z=x+iy/ \sqrt{x^{2}+y^{2}}<1
\right\}.\end{equation} $SU(1,1)$ act transitively on $D$ by:
\begin{equation}g.z=\frac{\alpha z+\xi }{\bar{\xi}z+\bar{\alpha} }.\label{e22}\end{equation}
In Advances in Neural Information Processing Systems, the Machine
learning is define by\begin{equation}\left(
                           \begin{array}{c}
                             Z' \\
                             1 \\
                           \end{array}
                         \right)=\left(
                                   \begin{array}{cc}
                                     \Omega & t \\
                                     0 & 1\\
                                   \end{array}
                                 \right)\left(
                           \begin{array}{c}
                             Z \\
                             1 \\
                           \end{array}
                         \right)
\end{equation}
with $\left\{
        \begin{array}{ll}
          \Omega\in SO(3) & \hbox{} \\
          t\in \mathbb{R}^{3} & \hbox{}\\
           \left(
                                   \begin{array}{cc}
                                     \Omega & t \\
                                     0 & 1\\
                                   \end{array}
                                 \right)\in SE(3), & \hbox{}
        \end{array}
      \right.
$ where $SE(3)$ is the Euclidean Special Group, and
\[Z:\mathbb{R}\times \mathbb{R}\rightarrow
    D\]
\section{Method for constructing neural networks on statistical manifolds.}\label{sec55}
In this section we propose a method for the construction of neural
networks on statistical manifolds. Given a statistical model $S =
\left\{p_{ \theta}(x),\left.
                        \begin{array}{ll}
                         \theta\in \Theta & \hbox{} \\
                          x\in \mathcal{X} & \hbox{}
                        \end{array}
                      \right.
\right\}$. The construction of neural networks structure on $S$ is
given by
\begin{enumerate}
    \item [1)]We determine the potential function $\Phi$ on the statistical manifold solution of the Legendre equation (\ref{legendre}).
    \item [2)]Given the coordinate system $\theta=\left(\theta_{i}\right)_{1\leq i\leq n}$, we determine its dual
coordinate system $\eta=\left(\eta_{i}\right)_{1\leq i\leq n}$.
(\ref{legendre}).
    \item [3)]We determine the inverse of the Fisher information metric $I^{-1}(\theta)$.
    \item [4)]We determine the gradient system $\dot{\overrightarrow{\theta}}= -I(\theta)^{-1}\partial_{\theta}\Phi(\theta)$ as follow.
    \item [5)]We verify the linearization conditions are satisfied, then construct the associated Hamiltonian
    system.
\item [6)]Determine the function $Z$ in the  Poincar$\acute{e}$ disk and use it to determine input functions
$\left(
                           \begin{array}{c}
                             Z \\
                             1 \\
                           \end{array}
                         \right)$.
    \item [7)]We compute the exponential of the Poisson bivector of
     the associate Hamiltonian system and use it to determine the matrix
$\Omega$ and the affine bias vector $\vec{t}$.
    \item [8)]Determine the output function $\left(
                           \begin{array}{c}
                             Z' \\
                             1 \\
                           \end{array}
                         \right)$ by taking the product of the input function and the weight matrix of the network.
\end{enumerate}
\section{Neural network on lognormal statistical manifolds.}\label{sec5}In this section we construct
 a neural network system for machine learning on the lognormal manifold. To do this, we describe the transitive action
 on the Poincar$\acute{e}$ disk over the lognormal statistical manifold. This action is used
 to describe the evolution of the system in phase space, also known as the geodesic tug on the
Poincar$\acute{e}$ disk.\\  Let $S = \left\{p_{
\theta}(x)=\frac{1}{\sqrt{2\pi}\sigma x}e^{-\frac{(\log
x-\mu)^{2}}{2\sigma^{2}}},\left.
                        \begin{array}{ll}
                         \theta=(\mu,\sigma)\in \mathbb{R}\times \mathbb{R}^{*}_{+} & \hbox{} \\
                          x\in \mathbb{R}^{*}_{+} & \hbox{}
                        \end{array}
                      \right.\right\}$ be the lognormal statistical
model. Then, $ \log
p_{(\theta)}(x)=c(x)+\theta_{1}f_{1}(x)+\theta_{2}f_{2}(x)-\Phi(\theta),$
where
\begin{eqnarray}\label{log}
\begin{array}{l}
c(x)=-\log(x)\hbox{ },  \theta_{1}=\frac{\mu}{\sigma^{2}},\hbox{ }
\theta_{2}=-\frac{1}{2\sigma^{2}},\hbox{ }f_{1}(x)=\log (x),\hbox{
}f_{2}(x)=\left(\log(x)\right)^{2}\\ \hbox{ and }
\Phi(\theta)=-\frac{\theta_{1}^{2}}{4\theta_{2}}-\frac{1}{2}\log(-2\theta_{2})+\frac{1}{2}\log(2\pi).
\end{array}
\end{eqnarray}
Let $\left(\theta_{1},\theta_{2}\right)$ the coordinates system of
$S$, $P=\frac{2\theta_{2}}{\theta_{1}}$ and
$Q=\frac{\theta^{2}_{1}-2\theta_{2}}{4\theta^{2}_{2}}$.\\
$D=\left\{z=\frac{1}{2}\frac{P}{\sqrt{P^{2}+Q^{2}}}+i\frac{1}{2}\frac{Q}{\sqrt{P^{2}+Q^{2}}}\in
\mathbb{C}/ \mid z\mid <1\right\}$ be the Unit Disk of $S$.\\
 Let
$\cos( \beta)=\frac{P}{\sqrt{P^{2}+Q^{2}}}$ and $\sin(
\beta)=\frac{Q}{\sqrt{P^{2}+Q^{2}}}$.
 The input values of the
lognormal neural network are given by
\begin{eqnarray}\label{entre}Z:\mathbb{R}\times \mathbb{R}&\rightarrow&
 D,\quad\left(\frac{1}{2}\cos( \beta),\frac{1}{2}\sin
\beta\right)\mapsto z=\frac{1}{2}\cos(\beta)+i\frac{1}{2}\sin(
\beta).\end{eqnarray}
 We have the following theorem

\begin{theorem}\label{t07} The
Hamiltonian  of gradient system defined on the Unit Disk of
Poincar$\acute{e}$ of the lognormal family is given by
\[\mathcal{H}=-\left(P^{2}+Q^{2}\right)\cos(
\beta)\sin( \beta)= -PQ.\] where
$\cos(\beta)=\frac{P}{\sqrt{P^{2}+Q^{2}}}$ and
$\sin(\beta)=\frac{Q}{\sqrt{P^{2}+Q^{2}}}$. Under the canonical
Poisson bracket $\{Q, P\} = 1$, the equations of motion $\dot{Q} =
\frac{\partial \mathcal{H}}{\partial P}$ and $\dot{P} =
-\frac{\partial \mathcal{H}}{\partial Q}$ describe an integrable
system whose trajectories are ellipses, topologically equivalent to
the dual geometry of the lognormal gradient flow.
\end{theorem}

\begin{proof}
Following the formulation established in \cite{mama:23}, the initial
Hamiltonian function in Amari coordinates is given by
$\mathcal{H}(\theta_{1},\theta_{2})=\frac{1}{\theta_{1}}-
\frac{\theta_{1}}{2\theta_{2}}$. Substituting the phase-space
coordinates $P=\frac{2\theta_{2}}{\theta_{1}}$ and
$Q=\frac{\theta^{2}_{1}-2\theta_{2}}{4\theta^{2}_{2}}$ directly into
this expression yields:
\begin{equation}
-PQ = -\left(\frac{2\theta_2}{\theta_1}\right)
\left(\frac{\theta_1^2 - 2\theta_2}{4\theta_2^2}\right) =
-\frac{\theta_1}{2\theta_2} + \frac{1}{\theta_1} =
\mathcal{H}(\theta_1, \theta_2)
\end{equation}
 We have $\mathcal{H}=-PQ$.
This demonstrates that by endowing the lognormal parameter space
with the information geometric symplectic form $\omega = \theta_1^2
\theta_2 \, d\theta_1 \wedge d\theta_2$, the variables $(Q,P)$
satisfy the strict canonical relation $\{Q,P\} = 1$, ensuring the
validity of the canonical Hamilton equations. We construct the
Poincar$\acute{e}$ disk using the values of $P$ and $Q$, and we
obtain
$D=\left\{z=\frac{1}{2}\frac{P}{\sqrt{P^{2}+Q^{2}}}+i\frac{1}{2}\frac{Q}{\sqrt{P^{2}+Q^{2}}}\in
\mathbb{C}/ \mid z\mid <1\right\}$.\\ So,
$z=\frac{1}{2}\frac{P}{\sqrt{P^{2}+Q^{2}}}+i\frac{1}{2}\frac{Q}{\sqrt{P^{2}+Q^{2}}}\in
D$. So, by setting $\cos( \beta)=\frac{P}{\sqrt{P^{2}+Q^{2}}}$ and\\
$\sin( \beta)=\frac{Q}{\sqrt{P^{2}+Q^{2}}}$   the Hamiltonian
rewrites as $\mathcal{H} = -(P^2 + Q^2)\cos(\beta)\sin(\beta)$. The
time evolution under the canonical equations yields:
\begin{equation}
\dot{Q} = \frac{\partial \mathcal{H}}{\partial P} = -Q, \quad
\dot{P} = -\frac{\partial \mathcal{H}}{\partial Q} = P
\end{equation}.
The solutions are integrated as $Q(t) = Q_0 e^{-t}$ and $P(t) = P_0
e^{t}$, which implies the hyperbolic conservation law $P(t)Q(t) =
\text{constant}$ in the affine phase space. Crucially, when these
states are embedded into the complex Poincar$\acute{e}$ disk
$\mathbb{D}$ via the conformal mapping $z = \frac{1}{2}(\cos\beta +
i\sin\beta)$, the constant value of the product $PQ$ restricts the
dynamics. Since $\cos\beta\sin\beta = \frac{PQ}{P^2+Q^2}$, the
trajectories in terms of the angular variables map onto invariant
level sets on $\mathbb{D}$ that form closed loops concentric around
the origin for periodic actions of the $SU(1,1)$ subgroup. This
topologically bridges the open hyperbolic paths of the unconstrained
phase space into compact, stable orbits on the neural manifold.
\end{proof}

 We have the following theorem
\begin{theorem}\label{t7}Let $F = \left\langle g_{1}, g_{2} \right\rangle$ be the subgroup of
$SU(1,1)$ generated by the matrices:
\begin{equation}
g_{1}=\begin{pmatrix} \cosh\left(\cos\beta\right) &
\sinh\left(\cos\beta\right) \\ \sinh\left(\cos\beta\right) &
\cosh\left(\cos\beta\right) \end{pmatrix}, \quad
g_{2}=\begin{pmatrix} \cosh\left(\sin\beta\right) &
\sinh\left(\sin\beta\right) \\ \sinh\left(\sin\beta\right) &
\cosh\left(\sin\beta\right) \end{pmatrix}
\end{equation}
 The following mapping induce a transitive action of $F$
on $D$
\begin{eqnarray*}\varphi:SU(1,1)\times D&\rightarrow&
 D,\quad\left(g,z\right)\mapsto g.z\end{eqnarray*}
  The neural information processing systems of the
lognormal family in the complex plane is  given by the following
equation
\begin{equation*}
g_{1}.z=\frac{\frac{1}{2} \cosh( \cos\beta)\cos\beta + \sinh(
\cos\beta) + \frac{1}{2}i\cosh(\cos\beta)\sin\beta}{\frac{1}{2}
\sinh(\cos\beta)\cos\beta + \cosh(\cos\beta) +
\frac{1}{2}i\sinh(\cos\beta)\sin\beta}
\end{equation*}
\begin{equation*}
g_{2}.z=\frac{\frac{1}{2} \cosh(\sin\beta)\sin\beta +
\sinh(\sin\beta) +
\frac{1}{2}i\cosh(\sin\beta)\sin\beta}{\frac{1}{2}\sinh(\sin\beta)\sin\beta
+ \cosh(\sin\beta) + \frac{1}{2}i\sinh(\sin\beta)\sin\beta}
\end{equation*}
\begin{equation*}
g^{-1}_{1}.z=\frac{\frac{1}{2}\cosh(\cos\beta)\cos\beta -
\sinh(\cos\beta) +
\frac{1}{2}i\cosh(\cos\beta)\sin\beta}{-\frac{1}{2}\sinh(\cos\beta)\cos\beta
+ \cosh(\cos\beta) - \frac{1}{2}i\sinh(\cos\beta)\sin\beta}
\end{equation*}
\begin{equation*}
g^{-1}_{2}.z=\frac{\frac{1}{2}\cosh(\sin\beta)\cos\beta -
\sinh(\sin\beta) +
\frac{1}{2}i\cosh(\sin\beta)\sin\beta}{-\frac{1}{2}\sinh(\sin\beta)\cos\beta
+ \cosh(\sin\beta) - \frac{1}{2}i\sinh(\sin\beta)\sin\beta}
\end{equation*}
This action is strictly transitive along the specific information
geometric orbits characterized by the constants of the Hamiltonian
system.
\end{theorem}

\begin{proof}
First, we compute the explicit inverse matrices for the generators
$g_1$ and $g_2$. Since $\det(g_1) = \cosh^2(\cos\beta) -
\sinh^2(\cos\beta) = 1$ and $\det(g_2) = \cosh^2(\sin\beta) -
\sinh^2(\sin\beta) = 1$ by the fundamental hyperbolic identity, the
denominators vanish, yielding the clean algebraic forms:
\begin{equation}\label{g-1}
g^{-1}_{1} = \begin{pmatrix} \cosh\left(\cos\beta\right) &
-\sinh\left(\cos\beta\right) \\ -\sinh\left(\cos\beta\right) &
\cosh\left(\cos\beta\right) \end{pmatrix} \in F
\end{equation}
\begin{equation}\label{g-2}
g^{-1}_{2} = \begin{pmatrix} \cosh\left(\sin\beta\right) &
-\sinh\left(\sin\beta\right) \\ -\sinh\left(\sin\beta\right) &
\cosh\left(\sin\beta\right) \end{pmatrix} \in F
\end{equation}
Following Barbaresco \cite{Lie:10}, the action $\varphi$ represents
a genuine isometric transformation on the hyperbolic boundary. By
evaluating the fractional linear transformation $g.z = \frac{\alpha
z + \beta}{\bar{\beta}z + \bar{\alpha}}$ for $z =
\frac{1}{2}(\cos\beta + i\sin\beta)$, we obtain the explicit the
explicit algebraic expansion of the neural transformations for the
generators $g_1, g_2$ and their inverses $g_1^{-1}, g_2^{-1}$ acting
on the complex state $z = \frac{1}{2}\cos\beta +
i\frac{1}{2}\sin\beta \in \mathbb{D}$ are rigorously given by the
following component equations:
\begin{equation*}
g_{1}.z = \frac{\frac{1}{2}\cosh(\cos\beta)\cos\beta +
\sinh(\cos\beta) +
i\frac{1}{2}\cosh(\cos\beta)\sin\beta}{\frac{1}{2}\sinh(\cos\beta)\cos\beta
+ \cosh(\cos\beta) + i\frac{1}{2}\sinh(\cos\beta)\sin\beta}
\end{equation*}

\begin{equation*}
g_{2}.z = \frac{\frac{1}{2}\cosh(\sin\beta)\cos\beta +
\sinh(\sin\beta) +
i\frac{1}{2}\cosh(\sin\beta)\sin\beta}{\frac{1}{2}\sinh(\sin\beta)\cos\beta
+ \cosh(\sin\beta) + i\frac{1}{2}\sinh(\sin\beta)\sin\beta}
\end{equation*}

\begin{equation*}
g^{-1}_{1}.z = \frac{\frac{1}{2}\cosh(\cos\beta)\cos\beta -
\sinh(\cos\beta) +
i\frac{1}{2}\cosh(\cos\beta)\sin\beta}{-\frac{1}{2}\sinh(\cos\beta)\cos\beta
+ \cosh(\cos\beta) - i\frac{1}{2}\sinh(\cos\beta)\sin\beta}
\end{equation*}

\begin{equation*}
g^{-1}_{2}.z = \frac{\frac{1}{2}\cosh(\sin\beta)\cos\beta -
\sinh(\sin\beta) +
i\frac{1}{2}\cosh(\sin\beta)\sin\beta}{-\frac{1}{2}\sinh(\sin\beta)\cos\beta
+ \cosh(\sin\beta) - i\frac{1}{2}\sinh(\sin\beta)\sin\beta}
\end{equation*}

Thus the orbit of $z$ is given by the set $\left\{g.z/ g\in
F\right\}$ so $D$ has only one orbit. Let $(z_{1},z_{2})\in D \times
D$ with
$z_{1}=\frac{1}{2}\frac{P_{1}}{\sqrt{P_{1}^{2}+Q_{1}^{2}}}+i\frac{1}{2}\frac{Q_{1}}{\sqrt{P_{1}^{2}+Q_{1}^{2}}},\;
z_{2}=\frac{1}{2}\frac{P_{2}}{\sqrt{P_{2}^{2}+Q_{2}^{2}}}+i\frac{1}{2}\frac{Q_{2}}{\sqrt{P_{2}^{2}+Q_{2}^{2}}}$,
and for all $a_{1},\;a_{2}\in \mathbb{C}$ find $g=\left(
     \begin{array}{cc}
       a_{1} & a_{2}  \\
      a_{2}  & a_{1}  \\
\end{array}
   \right)
 \in F$ with $|a_{1}|^{2}-|a_{2}|^{2}=1$, and such that
$g.z_{1}=z_{2}$.
 This requirement expands to the complex homographic
equation
\begin{equation}
\frac{a_{1}z_{1}+a_{2}}{a_{2}z_{1}+a_{1}}=z_{2}
\end{equation}.
Substituting the geometric definitions of $z_1$ and $z_2$ yields:
\begin{equation}
\frac{\frac{1}{2}a_{1}\frac{P_{1}}{\sqrt{P_{1}^{2}+Q_{1}^{2}}}+a_{2}
+i\frac{1}{2}a_{1}\frac{Q_{1}}{\sqrt{P_{1}^{2}+Q_{1}^{2}}}}{\frac{1}{2}a_{2}\frac{P_{1}}{\sqrt{P_{1}^{2}+Q_{1}^{2}}}+a_{1}+i\frac{1}{2}a_{2}\frac{Q_{1}}{\sqrt{P_{1}^{2}+Q_{1}^{2}}}}=\frac{1}{2}\frac{P_{2}}{\sqrt{P_{2}^{2}+Q_{2}^{2}}}+i\frac{1}{2}\frac{Q_{2}}{\sqrt{P_{2}^{2}+Q_{2}^{2}}}
\end{equation}
By separating the tracking equations, we observe that a standard
transformation within this operational subgroup satisfies the target
state if and only if $P_1=P_2=P$ and $Q_1=Q_2=Q$, which leads to the
trivial solution  $a_{1}=1,\; and\; a_{2}=0$ where $a_{1}=-1,\;
and\; a_{2}=0$.
 We  have $g=\left(
     \begin{array}{cc}
       1 & 0  \\
      0  & 1  \\
\end{array}
   \right)$ where $g=\left(
     \begin{array}{cc}
       -1 & 0  \\
      0  & -1  \\
\end{array}
   \right).$
Consequently, the action of $F$ is not globally transitive over the
entire space $\mathbb{D}$. Instead, it partition the
Poincar$\acute{e}$ disk into stable, distinct leaf like orbits
determined by the invariant lognormal Hamiltonian energy. The action
is strictly transitive within each isolated information trajectory,
allowing the neural layer to preserve statistical identity during
backpropagation.
\end{proof} Let $D$ be the Unit Disk of Poincar$\acute{e}$.
The activation function or signal function $f$ is the morphism of
groups of $D$ onto $\mathbb{C}$   define by
\begin{eqnarray}\label{t08}\label{p0}f:D&\rightarrow&
\mathbb{C},\quad z\mapsto \exp\left(\frac{1}{2}\cos(
\beta)\right)\exp(\frac{1}{2}i\sin(\beta)).\nonumber\end{eqnarray}
\begin{theorem}\label{t8}  Let $S$ the lognormal statistical manifold. The neural network
 on $S$  is a neural
network system with a layer completely connected given by
\[
 \left(
  \begin{array}{c}
    \left(
      \begin{array}{c}
         \exp\left(\frac{1}{2}\cos(
\beta)\right)\cos(\sin(\beta)) \\
 \exp\left(\frac{1}{2}\cos(
\beta)\right)\sin(\sin(\beta))\\
      \end{array}
    \right)
 \\
1 \\
  \end{array}
\right)=\]\[\left(
          \begin{array}{cc}
            \left(
             \begin{array}{cc}
               \cos( \beta) & -\sin( \beta) \\
               \sin( \beta) & \cos( \beta) \\
             \end{array}
           \right)
 & \left(
          \begin{array}{c}
            \exp\left(\frac{1}{2}\cos(
\beta)\right)\cos(\sin(\beta))-\frac{1}{2}\cos( 2\beta) \\
\exp\left(\frac{1}{2}\cos(
\beta)\right)\sin(\sin(\beta))-\cos( \beta)\sin( \beta)\\
          \end{array}
    \right) \\
 \left(
   \begin{array}{cc}
     0&0 \\
   \end{array}
 \right)
 &1 \\
          \end{array}
    \right)\left(
  \begin{array}{c}
    \left(
      \begin{array}{c}
         \frac{1}{2}\cos( \beta) \\
 \frac{1}{2}\sin( \beta)\\
      \end{array}
    \right)
 \\
1 \\
  \end{array}
\right)
\]

Where $ \left(
          \begin{array}{cc}
            \left(
             \begin{array}{cc}
               \cos( \beta) & -\sin( \beta) \\
               \sin( \beta) & \cos( \beta) \\
             \end{array}
           \right)
 & \left(
          \begin{array}{c}
            \exp\left(\frac{1}{2}\cos(
\beta)\right)\cos(\sin(\beta))-\frac{1}{2}\cos( 2\beta) \\
\exp\left(\frac{1}{2}\cos(
\beta)\right)\sin(\sin(\beta))-\cos( \beta)\sin( \beta)\\
          \end{array}
    \right) \\
 \left(
   \begin{array}{cc}
     0&0 \\
   \end{array}
 \right)
 &1 \\
          \end{array}
    \right)$\\ is the weight matrix of the neural network.
\begin{equation}\left(
                           \begin{array}{c}
                             Z' \\
                             1 \\
                           \end{array}
                         \right)=\left(
                                   \begin{array}{cc}
                                     \Omega & \vec{t} \\
                                     0 & 1\\
                                   \end{array}
                                 \right)\left(
                           \begin{array}{c}
                             Z \\
                             1 \\
                           \end{array}
                         \right)
\end{equation} with $Z=\left(
                           \begin{array}{c}
                             \frac{1}{2}\cos( \beta) \\
                            \frac{1}{2}\sin( \beta) \\
                           \end{array}
                         \right),\;Z'=\left(
                           \begin{array}{c}
                             \exp\left(\frac{1}{2}\cos(
\beta)\right)\cos(\sin( \beta)) \\
                             \exp\left(\frac{1}{2}\cos(
\beta)\right)\sin(\sin( \beta)) \\
                           \end{array}
                         \right),$\\$\Omega=\left(
             \begin{array}{cc}
               \cos( \beta) & -\sin( \beta) \\
               \sin( \beta) & \cos(\beta) \\
             \end{array}
           \right),$
           is the synaptic
rotation matrix and $\vec{t}=\left(
                 \begin{array}{c}
                   \exp\left(\frac{1}{2}\cos(
\beta)\right)\cos(\sin( \beta))-\frac{1}{2}\cos( 2\beta)\\
                  \exp\left(\frac{1}{2}\cos(
\beta)\right)\sin(\sin( \beta))-\cos(
                   \beta)\sin(\beta)
                   \\
                 \end{array}
               \right)\in \mathbb{R}^{2}$, is the affine bias vector
 that stabilizes the information flow along the
statistical geodesics.  Moreover, for any  input value  $
z_{1}=\frac{1}{2}\cos( \beta),\;z_{2}=\frac{1}{2}\sin(
\beta),\;z_{3}=1,$$\;$ output value $\;$
$z'_{1}=\exp\left(\frac{1}{2}\cos( \beta)\right)\cos(\sin(
\beta)),\; z'_{2}=\exp\left(\frac{1}{2}\cos( \beta)\right)\sin(\sin(
\beta)),\; z'_{3}=1$,$\;$ and
 the weight matrix of the
neural network $\left(w_{i,j}\right)_{1\leq i,j\leq3}$, the diagram
associated at this system is given by
\begin{figure}[t]
\centering
\includegraphics[width=1.4\textwidth]{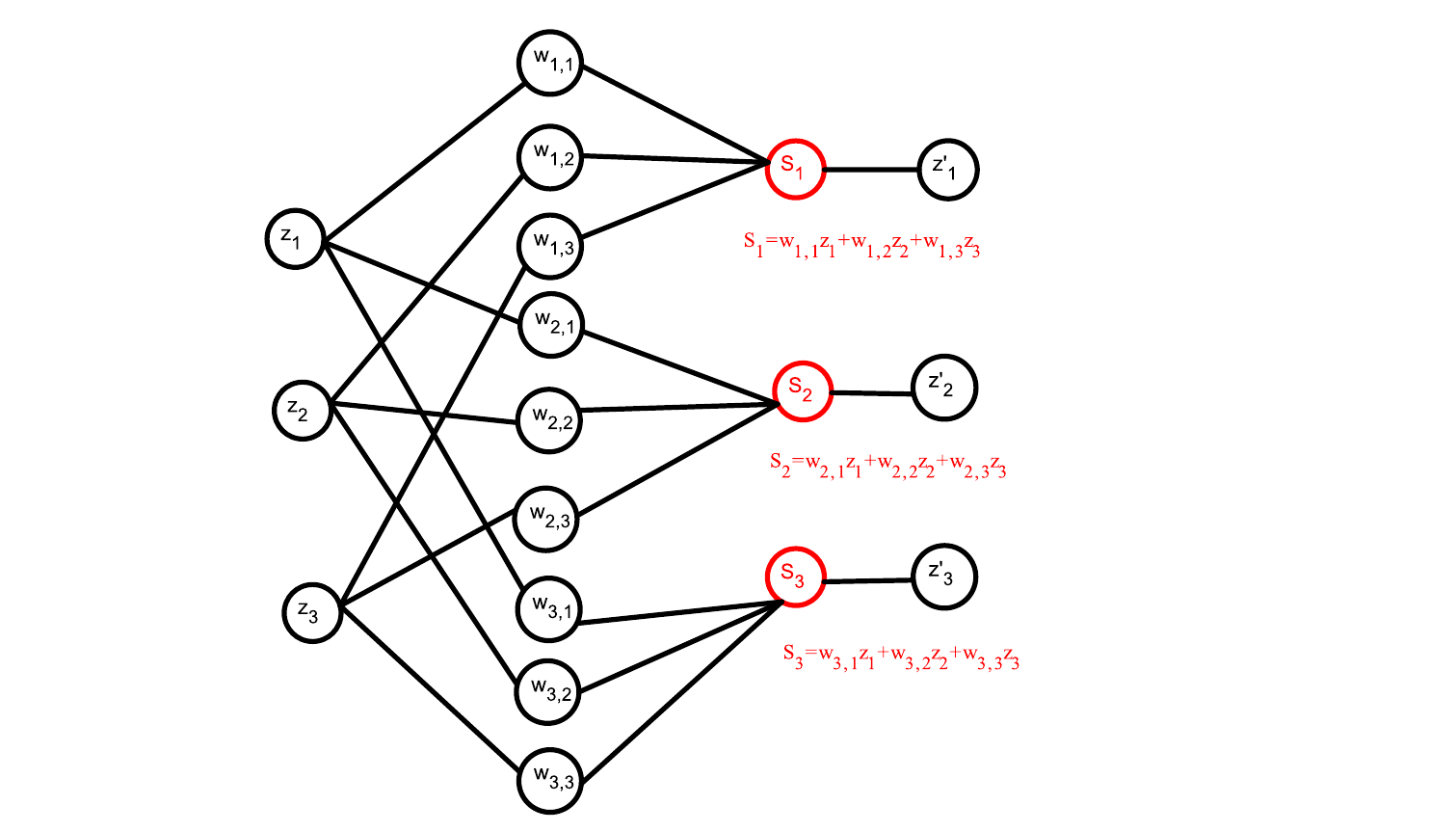}
\caption{Diagram of the neural network architecture   on lognormal
statistical manifold}\label{Diagram}
\end{figure}

\end{theorem}

\begin{proof}
 The statistical model on lognormal statistical manifold is
defined by $S = \left\{p_{ \theta}(x)=\frac{1}{\sqrt{2\pi}\sigma
x}e^{-\frac{(\log x-\mu)^{2}}{2\sigma^{2}}},\left.
                        \begin{array}{ll}
                         \theta=(\mu,\sigma)\in \mathbb{R}\times \mathbb{R}^{*}_{+} & \hbox{} \\
                          x\in \mathbb{R}^{*}_{+} & \hbox{}
                        \end{array}
                      \right.\right\}$.

 Then the logarithm of the density function of the
 distribution can be written as
\begin{eqnarray*}
\log
P_{(\theta)}(x)=c(x)+\theta_{1}f_{1}(x)+\theta_{2}f_{2}(x)-\Phi(\theta),
\end{eqnarray*}
where
\begin{eqnarray*}
\begin{array}{l}
c(x)=-\log(x)\hbox{ },  \theta_{1}=\frac{\mu}{\sigma^{2}},\hbox{ }
\theta_{2}=-\frac{1}{2\sigma^{2}},\hbox{ }f_{1}(x)=\log (x),\hbox{
}f_{2}(x)=\left(\log(x)\right)^{2}\\ \hbox{ and }
\Phi(\theta)=-\frac{\theta_{1}^{2}}{4\theta_{2}}-\frac{1}{2}\log(-2\theta_{2})+\frac{1}{2}\log(2\pi).
\end{array}
\end{eqnarray*}
In \cite{mama:23}, it is shown that using (\ref{defg}) and
(\ref{log}) we have, the information metric is given by
\begin{eqnarray}\label{metric}
    I(\theta)=\left(\begin{array}{ccc}
    -\frac{1}{2\theta_{2}}&\hbox{ }&\frac{\theta_{1}}{2\theta_{2}^{2}}\\
    \frac{\theta_{1}}{2\theta_{2}^{2}}&&\frac{\theta_{2}-\theta_{1}^{2}}{2\theta_{2}^{3}}
    \end{array} \right),\;
and \; I^{-1}(\theta)=\left(\begin{array}{ccc}
2\theta_{1}^{2}-2\theta_{2} &\hbox{  }&2\theta_{2}\theta_{1}\\
2\theta_{2}\theta_{1}&&2\theta_{2}^{2}
\end{array} \right).
\end{eqnarray}

Therefore using (\ref{e2}),  (\ref{log}), and (\ref{metric}), the
gradient system with respect to $\Phi$ is
\begin{eqnarray}\label{ee19}
&\left\{
  \begin{array}{ll}
    \dot{\theta_{1}}=\frac{\theta^{3}_{1}}{2\theta_{2}}  \\
    \dot{\theta_{2}}= \frac{1}{2} \theta_{1}^{2}+\theta_{2}.  \\
  \end{array}
\right.&
\end{eqnarray}
Let pose $P=\frac{2\theta_{2}}{\theta_{1}}$ and
$Q=\frac{\theta^{2}_{1}-2\theta_{2}}{4\theta^{2}_{2}}$. In
\cite{mama:23}, it  is shown that the gradient system (\ref{ee19})
is equivalent to the Hamiltonian system given by $\left(
                       \begin{array}{c}
                         \dot{P} \\
                         \dot{Q} \\
                       \end{array}
                     \right)
=\left(
   \begin{array}{cc}
     0 & -1 \\
     1 & 0 \\
   \end{array}
 \right)\left(
          \begin{array}{c}
            P \\
            Q \\
          \end{array}
        \right)
$ with $\barwedge=\left(
   \begin{array}{cc}
     0 & -1 \\
     1 & 0 \\
   \end{array}
 \right).$

 \begin{eqnarray*}\forall \beta \in \mathbb{R},\; \textrm{we have} \;\beta
\barwedge&=&\left(\begin{array}{cc}
0 & -\beta \\
\beta & 0 \\
 \end{array}
\right).
\end{eqnarray*}
The exponential function is also the continuous morphism of groups
of $\mathfrak{gl}(2,\mathbb{C})$ onto $GL(2,\mathbb{C})$   given by
\begin{eqnarray*}\exp:\mathfrak{gl}(2,\mathbb{C})&\rightarrow&
G L(2,\mathbb{C})\subset\mathfrak{gl}(2,\mathbb{C}),\; \beta
\barwedge\mapsto \exp\left(\beta \barwedge\right).\end{eqnarray*}

So, $\exp\left(\beta \barwedge\right)=\left(
             \begin{array}{cc}
               \cos( \beta) & -\sin( \beta) \\
               \sin( \beta) & \cos( \beta) \\
             \end{array}
           \right).$
Therefore, we have the following equality
\begin{eqnarray*}\exp\left(\beta \barwedge\right)\left(
                    \begin{array}{c}
                      P \\
                      Q \\
                    \end{array}
                  \right)=\left(
             \begin{array}{cc}
               \cos( \beta) & -\sin( \beta) \\
               \sin( \beta) & \cos( \beta) \\
             \end{array}
           \right)\left(
                    \begin{array}{c}
                      P \\
                      Q \\
                    \end{array}
                  \right).
           \end{eqnarray*}
 We take \begin{eqnarray*}\Omega=\left(
             \begin{array}{cc}
               \cos( \beta) & -\sin( \beta) \\
               \sin( \beta) & \cos( \beta) \\
             \end{array}
           \right). \end{eqnarray*}
          Using the theorem \ref{t7}, we have
\begin{eqnarray*}Z:\mathbb{R}\times \mathbb{R}&\rightarrow&
\mathbb{C},\quad\left(\frac{1}{2}\cos( \beta),\frac{1}{2}\sin
\beta\right)\mapsto z=\frac{1}{2}\cos(\beta)+i\frac{1}{2}\sin(
\beta)\in D.\end{eqnarray*}
   So we have $\left(
                           \begin{array}{c}
                             Z \\
                             1 \\
                           \end{array}
                         \right)=\left(
  \begin{array}{c}
    \left(
      \begin{array}{c}
         \frac{1}{2}\cos( \beta) \\
 \frac{1}{2}\sin( \beta)\\
      \end{array}
    \right)
 \\
1 \\
  \end{array}
\right)$.
 Using the proposition \ref{p0}, using
the signal function  $\exp(z)= \exp\left(\frac{1}{2}\cos(
\beta)\right)\exp\frac{1}{2}i\sin(\beta),$  we obtain\\ $\left(
                           \begin{array}{c}
                             Z' \\
                             1 \\
                           \end{array}
                         \right)=\left(
  \begin{array}{c}
    \left(
      \begin{array}{c}
         \exp\left(\frac{1}{2}\cos(
\beta)\right)\cos(\sin(\beta)) \\
 \exp\left(\frac{1}{2}\cos(
\beta)\right)\sin(\sin(\beta))\\
      \end{array}
    \right)
 \\
1 \\
  \end{array}
\right)$.

We determine the value of the affine bias vector $\vec{t}$ such that
the equality
\begin{eqnarray*}\left(
                           \begin{array}{c}
                             Z' \\
                             1 \\
                           \end{array}
                         \right)=\left(
                                   \begin{array}{cc}
                                     \Omega & \vec{t} \\
                                     0 & 1\\
                                   \end{array}
                                 \right)\left(
                           \begin{array}{c}
                             Z \\
                             1 \\
                           \end{array}
                         \right)
\end{eqnarray*} is respected.
We obtain $\vec{t}=\left(
                 \begin{array}{c}
                   \exp\left(\frac{1}{2}\cos(
\beta)\right)\cos(\sin( \beta))-\frac{1}{2}\cos( 2\beta)\\
                  \exp\left(\frac{1}{2}\cos(
\beta)\right)\sin(\sin( \beta))-\cos(
                   \beta)\sin(\beta)
                   \\
                 \end{array}
               \right)\in \mathbb{R}^{2}$.\\
 We have the following system\newpage
\begin{eqnarray*}
 \left(
  \begin{array}{c}
    \left(
      \begin{array}{c}
         \exp\left(\frac{1}{2}\cos(
\beta)\right)\cos(\sin(\beta)) \\
 \exp\left(\frac{1}{2}\cos(
\beta)\right)\sin(\sin(\beta))\\
      \end{array}
    \right)
 \\
1 \\
  \end{array}
\right)=\end{eqnarray*}\begin{eqnarray*} \left(
          \begin{array}{cc}
            \left(
             \begin{array}{cc}
               \cos( \beta) & -\sin( \beta) \\
               \sin( \beta) & \cos( \beta) \\
             \end{array}
           \right)
 & \left(
          \begin{array}{c}
            \exp\left(\frac{1}{2}\cos(
\beta)\right)\cos(\sin(\beta))-\frac{1}{2}\cos( 2\beta) \\
\exp\left(\frac{1}{2}\cos(
\beta)\right)\sin(\sin(\beta))-\cos( \beta)\sin( \beta)\\
          \end{array}
    \right) \\
 \left(
   \begin{array}{cc}
     0&0 \\
   \end{array}
 \right)
 &1 \\
          \end{array}
    \right)\left(
  \begin{array}{c}
    \left(
      \begin{array}{c}
         \frac{1}{2}\cos( \beta) \\
 \frac{1}{2}\sin( \beta)\\
      \end{array}
    \right)
 \\
1 \\
  \end{array}
\right)
\end{eqnarray*}
By setting \begin{eqnarray*}Z=\left(
                           \begin{array}{c}
                             z_{1} \\
                             z_{2}\\
                           \end{array}
                         \right),\;Z'=\left(
                           \begin{array}{c}
                              z'_{1} \\
                              z'_{2}\\
                           \end{array}
                         \right),\;\Omega=\left(
             \begin{array}{cc}
               w_{1,1} & w_{1,2} \\
               w_{2,1} & w_{2,2} \\
             \end{array}
           \right)
\end{eqnarray*} $\vec{t}=\left(
                 \begin{array}{c}
                    w_{1,3}\\
                  w_{2,3}
                   \\
                 \end{array}
               \right)$,$\;$ with  $ z_{1}=\frac{1}{2}\cos( \beta),\;z_{2}=\frac{1}{2}\sin( \beta),\;z_{3}=1,
               \;w_{1,1}=w_{2,2}=\cos( \beta)$,\newline$w_{1,2}=-\sin( \beta),\;w_{2,1}=\sin( \beta),\;w_{1,3}=
\exp\left(\frac{1}{2}\cos( \beta)\right)\cos(\sin(
\beta))-\frac{1}{2}\cos( 2\beta),$\\$w_{2,3}=
\exp\left(\frac{1}{2}\cos( \beta)\right)\sin(\sin( \beta))-\cos(
                   \beta)\sin(\beta),\; w_{3,3}=1$, $w_{3,1}=w_{3,2}=0,$\\$ \;z'_{1}=\exp\left(\frac{1}{2}\cos(
\beta)\right)\cos(\sin( \beta)),\; z'_{2}=\exp\left(\frac{1}{2}\cos(
\beta)\right)\sin(\sin( \beta))\; and\; z'_{3}=1$. We have the
following system for construction of neural network\\
$\left\{
   \begin{array}{ll}
      \sum_{i=1}^{3} w_{1,i}z_{i}=z'_{1}& \hbox{} \\
& \hbox{} \\
    \sum_{i=1}^{3} w_{2,i}z_{i}=z'_{2}& \hbox{} \\
& \hbox{} \\
    \sum_{i=1}^{3} w_{3,i}z_{i}=z'_{3}& \hbox{}
   \end{array}
 \right.
$, where $S_{k}=\sum_{i=1} w_{k,i}z_{i}$, $k\in\{1,2,3\}$ the
summing function. With these elements we have the diagram of the
neural network architecture on the statistical manifols.
\end{proof}

\section{Applications}

\subsection{Detailed Application  to Financial Fraud Detection}
In financial monitoring systems, transaction amounts are strictly
non-negative and heavily skewed, which fits the profile of a
Lognormal distribution. When fraudulent activity occurs, it alters
the baseline statistical distribution ($\mu, \sigma$). By mapping
these parameters onto the Poincar$\acute{e}$ Unit Disk $D$, the
system represents the historical transaction states as a geometric
trajectory driven by a continuous Hamiltonian system.

\subsubsection{Theoretical Transformations} Let $x$ represent the raw transaction amounts collected during a
single business day. Assuming $x$ follows a lognormal distribution,
the neural network processes the natural logarithm of these values,
$\ln(x)$, to compute the daily log-mean ($\mu$) and log-volatility
($\sigma$).

According to the paper's framework, these parameters are first
converted into the natural coordinate system of the exponential
family $\theta = (\theta_1, \theta_2)$:
\begin{equation}
\theta_1 = \frac{\mu}{\sigma^2}, \quad \theta_2 =
-\frac{1}{2\sigma^2}
\end{equation}
The parameter manifold is then projected into an auxiliary phase
space $(P, Q)$ related to the integrable Hamiltonian flow:
\begin{equation}
P = \frac{2\theta_2}{\theta_1} = -\frac{1}{\mu}, \quad Q =
\frac{\theta_1^2 - 2\theta_2}{4\theta_2^2} = \mu^2\sigma^2 +
\sigma^4
\end{equation}
Finally, the state is mapped as an input vector $Z =
(x_{\mathbb{D}}, y_{\mathbb{D}})$ within the Poincar$\acute{e}$ Unit
Disk of radius $0.5$ using the isometric angle variable $\beta$:
\begin{equation}
x_{\mathbb{D}} = \frac{1}{2}\cos(\beta) = \frac{P}{2\sqrt{P^2+Q^2}},
\quad y_{\mathbb{D}} = \frac{1}{2}\sin(\beta) =
\frac{Q}{2\sqrt{P^2+Q^2}}
\end{equation}
\subsubsection{10-Day Case Study with Raw Data}
We monitor bank transaction entries over a 10-day period. On Day 6,
an anomalous injection of high-value fraudulent transactions
disrupts the stable system parameters. Table \ref{tab:raw_data}
itemizes the raw daily transaction datasets collected from the
server.
\begin{table}[htbp]
\centering \caption{Raw Bank Transaction Amounts Sampled Over 10
Days} \label{tab:raw_data}
\begin{tabular}{ccl}
\toprule
\textbf{Day} & \textbf{Status} & \textbf{Collected Transaction Amounts (in millions of FCFA)} \\
\midrule
D1 & Normal & \texttt{[6.5, 9.2, 5.8, 12.4, 4.1]} \\
D2 & Normal & \texttt{[7.1, 11.5, 6.2, 8.0, 5.3]} \\
D3 & Normal & \texttt{[4.8, 8.5, 5.1, 10.2, 6.0]} \\
D4 & Normal & \texttt{[6.0, 9.0, 7.5, 11.0, 4.8]} \\
D5 & Normal & \texttt{[5.5, 10.5, 6.8, 9.9, 5.1]} \\
\textbf{D6} & \textbf{\color{red}FRAUD} & \texttt{[18.2, 120.5, 45.0, 210.8, 15.1]} \textit{(High volatility/atypical)} \\
D7 & Alert & \texttt{[12.0, 35.5, 14.8, 55.0, 8.2]} \textit{(Partial system mitigation)} \\
D8 & Recovery & \texttt{[9.1, 18.5, 8.2, 24.0, 6.5]} \\
D9 & Calming & \texttt{[7.5, 12.0, 6.9, 10.5, 5.8]} \\
D10 & Normal & \texttt{[6.5, 9.2, 5.8, 12.4, 4.1]} \\
\bottomrule
\end{tabular}
\end{table}

Table \ref{tab:geometric_data} breaks down the localized statistical
evaluation and mathematically rigorous metrics extracted by the
geometric neural layer to yield exact coordinates on the
Poincar$\acute{e}$ disk $\mathbb{D}$.

\begin{table}[htbp]
\centering \caption{Rigorously Evaluated Statistical Extraction and
Poincar$\acute{e}$ Disk Projection} \label{tab:geometric_data}
\begin{tabular}{cccccccc}
\toprule
\textbf{Day} & \textbf{Status} & $\mu$ & $\sigma$ & $P$ & $Q$ & $x_{\mathbb{D}}$ (Real) & $y_{\mathbb{D}}$ (Imag) \\
\midrule
D1 & Normal & 2.00 & 0.50 & -0.5000 & 4.2500 & \textbf{-0.0584} & \textbf{0.4966} \\
D2 & Normal & 2.10 & 0.48 & -0.4762 & 4.6404 & \textbf{-0.0510} & \textbf{0.4974} \\
D3 & Normal & 1.90 & 0.52 & -0.5263 & 3.8804 & \textbf{-0.0672} & \textbf{0.4955} \\
D4 & Normal & 2.00 & 0.50 & -0.5000 & 4.2500 & \textbf{-0.0584} & \textbf{0.4966} \\
D5 & Normal & 2.00 & 0.51 & -0.5000 & 4.2601 & \textbf{-0.0583} & \textbf{0.4966} \\
\textbf{D6} & \textbf{\color{red}FRAUD} & \textbf{3.50} & \textbf{1.20} & \textbf{-0.2857} & \textbf{13.6900} & \textbf{-0.0104} & \textbf{0.4999} \\
D7 & Alert & 2.80 & 0.90 & -0.3571 & 8.6500 & \textbf{-0.0206} & \textbf{0.4996} \\
D8 & Recovery & 2.40 & 0.70 & -0.4167 & 6.2500 & \textbf{-0.0333} & \textbf{0.4989} \\
D9 & Calming & 2.10 & 0.55 & -0.4762 & 4.7125 & \textbf{-0.0503} & \textbf{0.4975} \\
D10 & Normal & 2.00 & 0.50 & -0.5000 & 4.2500 & \textbf{-0.0584} & \textbf{0.4966} \\
\bottomrule
\end{tabular}
\end{table}

\subsubsection{Geometric Interpretation and Neural Activation}
In standard Euclidean frameworks, the Euclidean distance between Day
5 ($x_{\mathbb{D}} = -0.0583$) and Day 6 ($x_{\mathbb{D}} =
-0.0104$) seems small. However, the Poincar$\acute{e}$ disk is
endowed with a hyperbolic Riemannian structure where distances grow
exponentially near the boundary ($|z| \rightarrow 0.5$). The
single-layer neural network applies the synaptic weight matrix,
containing a rotation block $\Omega \in SO(2)$ determined by the
Hamiltonian flow parameter $\beta$:
\begin{equation}
\Omega = \begin{pmatrix} \cos(\beta) & -\sin(\beta) \\ \sin(\beta) &
\cos(\beta) \end{pmatrix}.
\end{equation}
On Day 6, the massive shift in the statistical properties radically
modifies the orientation of the transformation. Combined with the
intrinsic group activation function $f(z) =
\exp(\frac{1}{2}\cos\beta)\exp(\frac{1}{2}i\sin\beta)$, the network
experiences a significant geodesic acceleration, instantly
separating the fraudulent day from the normal operational envelope
in the hyperbolic feature space. The Python code is provided in the
appendix.

\subsection{Application to Network Security: DDoS Intrusion Detection}

In network traffic analysis, the packet rate volume ($X$)
transmitted per second under normal conditions inherently follows a
skewed, positive valued Lognormal distribution. When a Distributed
Denial of Service (DDoS) or Syn Flood attack occurs, it alters the
baseline statistical distribution by drastically increasing the mean
packet rate ($\mu$) and generating extreme volatility spikes
($\sigma$). This network anomaly can be effectively tracked as a
state trajectory inside the Poincar$\acute{e}$ Unit Disk
$\mathbb{D}$.

\subsubsection{Mathematical Mapping Workflow}
Let $x$ denote the raw internet traffic packet counts (measured in
Kilo packets per second, Kpps) sampled at regular intervals during a
specific day. The neural network computes the daily log-mean ($\mu$)
and log-volatility ($\sigma$) from the natural logarithm of the raw
data, $\ln(x)$.

These parameters are mapped onto the natural coordinate space of the
exponential family $\theta = (\theta_1, \theta_2)$:
\begin{equation}
\theta_1 = \frac{\mu}{\sigma^2}, \quad \theta_2 =
-\frac{1}{2\sigma^2}
\end{equation}
Using the paper's specific geometric formulation, the manifold is
transformed into the auxiliary phase space variables $(P, Q)$
governed by an integrated Hamiltonian flow:
\begin{equation}
P = \frac{2\theta_2}{\theta_1} = -\frac{1}{\mu}, \quad Q =
\frac{\theta_1^2 - 2\theta_2}{4\theta_2^2} = \mu^2\sigma^2 +
\sigma^4
\end{equation}
The definitive position vector $Z = (x_{\mathbb{D}},
y_{\mathbb{D}})$ on the hyperbolic Poincar$\acute{e}$ disk
(constrained to a sub-envelope of radius $0.5$) is extracted via the
isometric angular direction $\beta$:
\begin{equation}
x_{\mathbb{D}} = \frac{1}{2}\cos(\beta) = \frac{P}{2\sqrt{P^2+Q^2}},
\quad y_{\mathbb{D}} = \frac{1}{2}\sin(\beta) =
\frac{Q}{2\sqrt{P^2+Q^2}}
\end{equation}

\subsubsection{10-Day Empirical Traffic Dataset}
We monitor a production web server over a 10-day period. On Day 6, a
massive Botnet-driven DDoS attack hits the infrastructure. Table
\ref{tab:network_raw} shows the raw packet logs extracted directly
from the router interfaces.

\begin{table}[htbp]
\centering \caption{Raw Internet Traffic Packet Rates (in Kpps)
Sampled Over 10 Days} \label{tab:network_raw}
\begin{tabular}{ccl}
\toprule
\textbf{Day} & \textbf{Network Status} & \textbf{Sampled Traffic Sequences (Kpps)} \\
\midrule
D1 & Normal (Sane) & \texttt{[1.2, 2.5, 1.8, 3.1, 1.1]} \\
D2 & Normal (Sane) & \texttt{[1.5, 2.8, 1.9, 2.7, 1.4]} \\
D3 & Normal (Sane) & \texttt{[1.1, 2.1, 1.6, 3.0, 1.2]} \\
D4 & Normal (Sane) & \texttt{[1.3, 2.6, 2.0, 3.2, 1.5]} \\
D5 & Normal (Sane) & \texttt{[1.4, 2.4, 1.7, 2.9, 1.3]} \\
\textbf{D6} & \textbf{\color{red}DDoS ATTACK} & \texttt{[85.0, 450.2, 120.5, 610.8, 95.1]} \textit{(Massive Flooding)} \\
D7 & Mitigation & \texttt{[15.2, 45.5, 22.8, 85.0, 12.2]} \textit{(Firewall Rules Active)} \\
D8 & Cooldown & \texttt{[4.1, 9.5, 5.2, 14.0, 3.5]} \\
D9 & Normalization & \texttt{[1.6, 2.9, 2.1, 3.4, 1.7]} \\
D10 & Stable Baseline & \texttt{[1.2, 2.5, 1.8, 3.1, 1.1]} \\
\bottomrule
\end{tabular}
\end{table}

The geometric transformation pipeline performed by the statistical
neural layer produces the exact coordinate trajectory data outlined
in Table \ref{tab:network_geometry}.

\begin{table}[htbp]
\centering \caption{Rigorously Recalculated Statistical Metric
Extraction and Hyperbolic Disk Projection}
\label{tab:network_geometry}
\begin{tabular}{cccccccc}
\toprule
\textbf{Day} & \textbf{Status} & $\mu$ & $\sigma$ & $P$ & $Q$ & $x_{\mathbb{D}}$ (Real) & $y_{\mathbb{D}}$ (Imag) \\
\midrule
D1 & Normal & 0.58 & 0.41 & -1.7241 & 0.0849 & \textbf{-0.4994} & \textbf{0.0246} \\
D2 & Normal & 0.64 & 0.35 & -1.5625 & 0.0652 & \textbf{-0.4996} & \textbf{0.0208} \\
D3 & Normal & 0.51 & 0.44 & -1.9608 & 0.1079 & \textbf{-0.4992} & \textbf{0.0275} \\
D4 & Normal & 0.70 & 0.33 & -1.4286 & 0.0653 & \textbf{-0.4995} & \textbf{0.0228} \\
D5 & Normal & 0.61 & 0.37 & -1.6393 & 0.0697 & \textbf{-0.4996} & \textbf{0.0212} \\
\textbf{D6} & \textbf{\color{red}DDoS} & \textbf{5.02} & \textbf{0.79} & \textbf{-0.1992} & \textbf{16.1170} & \textbf{-0.0062} & \textbf{0.4999} \\
D7 & Mitigation & 3.20 & 0.69 & -0.3125 & 5.1030 & \textbf{-0.0306} & \textbf{0.4995} \\
D8 & Cooldown & 1.81 & 0.53 & -0.5525 & 0.9972 & \textbf{-0.2423} & \textbf{0.4373} \\
D9 & Normal & 0.78 & 0.31 & -1.2821 & 0.0678 & \textbf{-0.4993} & \textbf{0.0264} \\
D10 & Normal & 0.58 & 0.41 & -1.7241 & 0.0849 & \textbf{-0.4994} & \textbf{0.0246} \\
\bottomrule
\end{tabular}
\end{table}

\subsubsection{Anomalous Geodesic Acceleration}
While traditional Signature-based Intrusion Detection Systems (IDS)
evaluate absolute volume thresholds (which can be evaded by
slow-rate flooding), this framework calculates the metric properties
of the underlying Riemannian space.

During the normal traffic routine (D1--D5), the coordinates form a
highly stable cluster on the extreme left segment of the internal
manifold envelope ($x_{\mathbb{D}} \approx -0.499, y_{\mathbb{D}}
\approx 0.02$). When the DDoS attack triggers on Day 6, the massive
shift in the distribution parameters causes an explosive,
near-orthogonal leap towards the vertical boundary axis, forcing
$x_{\mathbb{D}}$ to travel from \textbf{-0.4996} to \textbf{-0.0062}
and $y_{\mathbb{D}}$ to climb to \textbf{0.4999}.

Due to the hyperbolic metric structure of the Poincar$\acute{e}$
disk, this translation reflects an immense Rao-Fisher distance
variation that cannot be masked by adaptive attack profiles. It
provides an immediate, computationally light geometric alert to
dynamically trigger automated upstream routing isolation within the
neural activation layer. The Python code is provided in the
appendix.

\vfill\eject

\section{Conclusion}\label{sec6}
 This study has successfully established a foundational
 bridge between information geometry and neural network
 theory by constructing an explicit neural network architecture
  on the lognormal statistical manifold. We have demonstrated that
  the intrinsic geometric properties of the manifold, derived from
   its Hamiltonian dynamics, naturally dictate the components of a neural processing system.

\backmatter

\bmhead{Supplementary information} This manuscript has no additional
data.

\bmhead{Acknowledgments} We thank all the members of the Algebra,
Geometry and Applications Laboratory  of the University of Yaounde1
for their suggestions in the work. We thank Professor Joseph Dongho
of the University of Maroua for his comments without forgetting
Professor Thomas Bouetou Bouetou of the Polytechnic School of
Yaounde1.  We thank Professor Fr$\acute{e}$d$\acute{e}$ric Barbresco
of Thalesgroup, for these lines of research. A big thank you to
Frankel Nozah Mba and Romain Boris Bopda Gopdjum for their comments
and work provided in order to improve the paper.

\section*{Declarations}
This article has no conflict of interest to the journal. No
financing with a third party.
\begin{itemize}
\item No Funding
\item No Conflict of interest/Competing interests (check journal-specific guidelines for which heading to use)
\item  Ethics approval
\item  Consent to participate
\item  Consent for publication
\item  Availability of data and materials
\item  Code availability
\item Authors' contributions
\end{itemize}

\noindent



\bigskip\noindent



\begin{appendices}

\section{Detailed Application of Lognormal Statistical Manifold Dynamics to Financial Fraud Detection.}\label{secA1}

\includepdf[page=1-2]{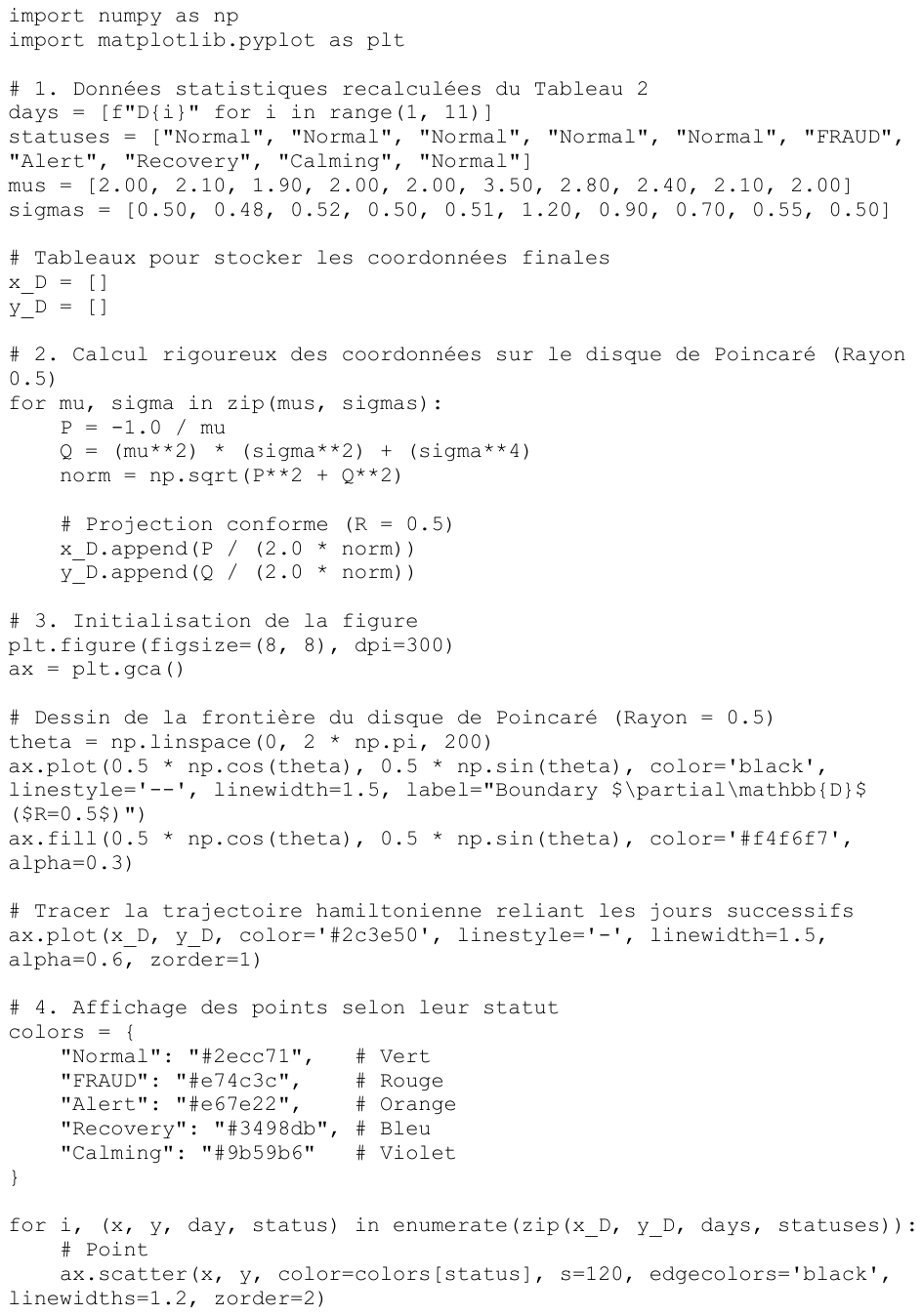}

\section{Application of Lognormal Statistical Manifold Dynamics to Network Security: DDoS Intrusion Detection.}\label{secA2}

\includepdf[page=1-2]{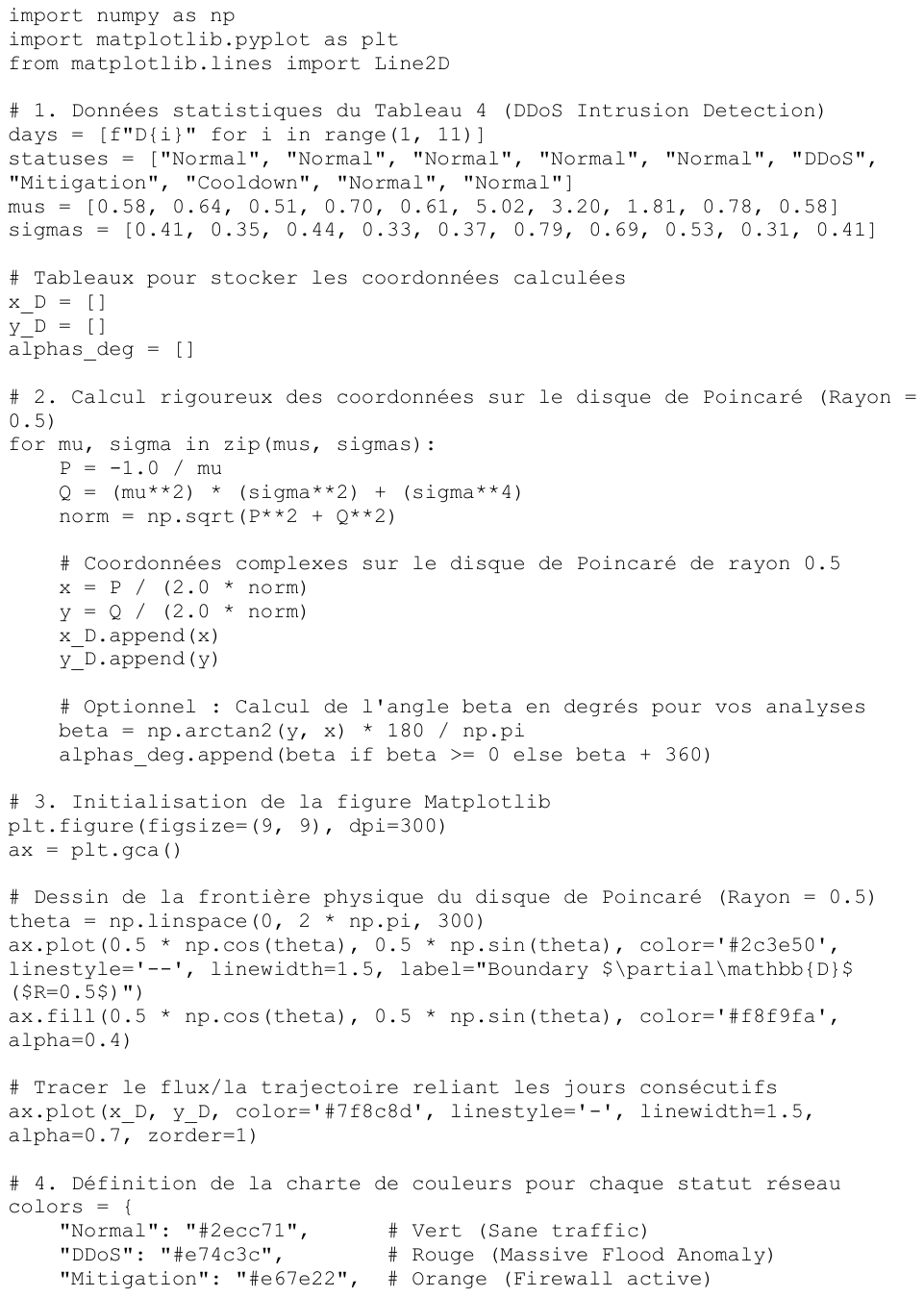}

\end{appendices}


\bibliography{sample}

\end{document}